\documentclass[runningheads]{llncs}

\usepackage{eccv}

\usepackage{eccvabbrv}

\usepackage{graphicx}
\usepackage{booktabs}
\usepackage{wrapfig}
\usepackage{epigraph}
\usepackage{multirow}
\usepackage{multicol}
\usepackage{colortbl}
\definecolor{cvprblue}{rgb}{0.21,0.49,0.74}
\definecolor{citecolor}{HTML}{229954}
\definecolor{mygray}{gray}{.9}
\usepackage[accsupp]{axessibility}  

\usepackage[pagebackref,breaklinks,colorlinks,citecolor=eccvblue]{hyperref}

\usepackage{orcidlink}

\begin{document}

\title{Learning Long-form Video Prior via Generative Pre-Training} 

\titlerunning{Long-form Video Prior}

\author{Jinheng Xie\inst{1} \,
Jiajun Feng\inst{1*} \,
Zhaoxu Tian\inst{1*} \, Kevin Qinghong Lin\inst{1} \\ Yawen Huang\inst{2}$^{\dagger}$\, Xi Xia\inst{1}\, Nanxu Gong\inst{1} \,  Xu Zuo\inst{1} \, Jiaqi Yang\inst{1} \\ Yefeng Zheng\inst{2} \, Mike Zheng Shou\inst{1}$^{\dagger}$}

\authorrunning{Xie et al.}
{\let\thefootnote\relax\footnotetext{$^{\dagger}$ Corresponding Author}}
{\let\thefootnote\relax\footnotetext{$^{*}$ Equal Contribution}}
\institute{Show Lab, National University of Singapore \and
Jarvis Research Center, Tencent Youtu Lab
\\
\email{\{sierkinhane, mike.zheng.shou\}@gmail.com}}
\maketitle
\begin{abstract}
  Concepts involved in long-form videos such as people, objects, and their interactions, can be viewed as following an implicit prior. They are notably complex and continue to pose challenges to be comprehensively learned. In recent years, generative pre-training (GPT) has exhibited versatile capacities in modeling any kind of text content even visual locations. Can this manner work for learning long-form video prior? Instead of operating on pixel space, it is efficient to employ visual locations like bounding boxes and keypoints to represent key information in videos, which can be simply discretized and then tokenized for consumption by GPT. Due to the scarcity of suitable data, we create a new dataset called \textbf{Storyboard20K} from movies to serve as a representative. It includes synopses, shot-by-shot keyframes, and fine-grained annotations of film sets and characters with consistent IDs, bounding boxes, and whole body keypoints. In this way, long-form videos can be represented by a set of tokens and be learned via generative pre-training. Experimental results validate that our approach has great potential for learning long-form video prior. Code and data will be released at \url{https://github.com/showlab/Long-form-Video-Prior}.
\end{abstract}

\section{Introduction}
\label{sec:intro}

Continuously captured by digital cameras, our world, spanning from the past to the present and into the future, can be viewed as long-form videos. Its continuous stream of moving pictures unfolds a comprehensive narrative involving people, objects, and their interactions. These kinds of elements together can be viewed as following \textit{an implicit prior}, which is extremely hard to formulate and poses challenges to be learned by current state-of-the-art approaches. For instance, though recent diffusion models~\cite{sohl2015deep,ho2020denoising,sd} have ushered in a new era of video generation~\cite{video1, videofusion, make-a-video}, they continue to face challenges in generating short videos and are not well-suited for long-form scenarios. This is attributed to the extended temporal coverage, substantial variations across frames, and intricate interactions among consistent characters in long-form videos.

\begin{figure*}[t]
    \centering
    \includegraphics[width=\linewidth]{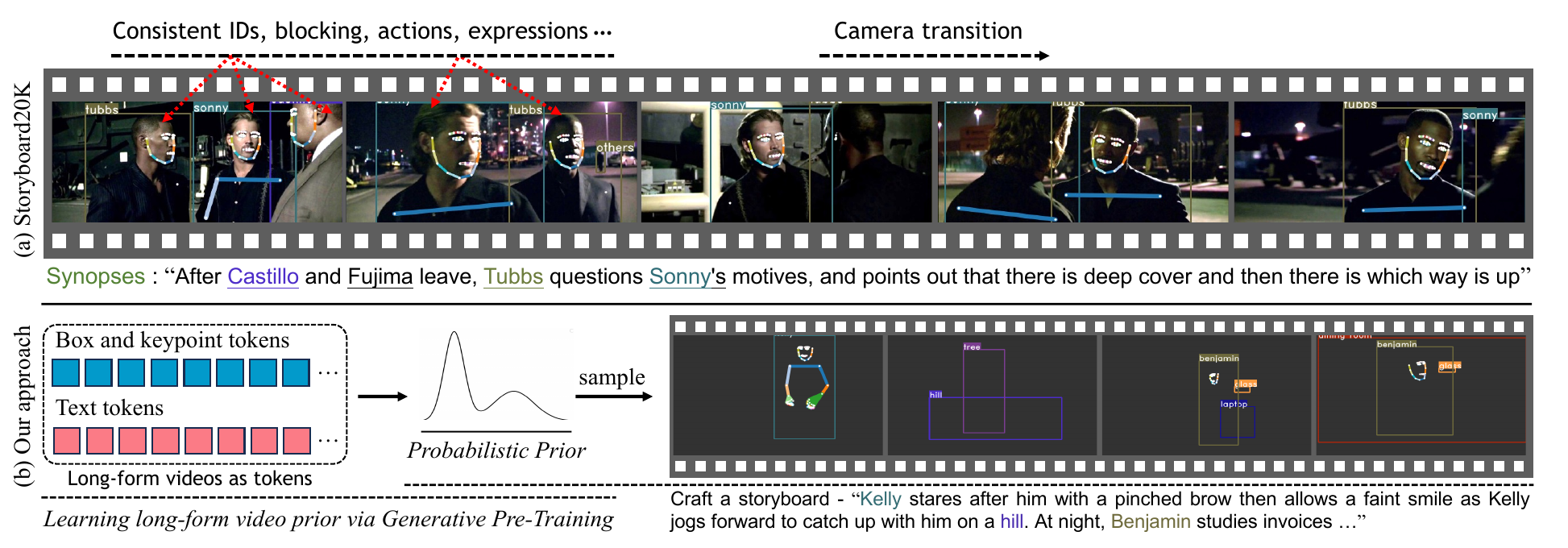}
    \captionof{figure}{(a) Samples from the proposed \textbf{Storyboard20K}. It consists of scripts, shot-by-shot keyframes, and fine-grained annotations (bounding boxes and whole body keypoints) of characters and film sets. (b) The proposed approach. Instead of modeling in pixel space, we propose to represent movies as sequences of tokens that can be jointly learned with the script via generative pre-training.}
    \label{fig:teaser}
    \vspace{-2pt}
\end{figure*}%

Over the past few years, generative pre-trained transformers (GPTs)~\cite{gpt2, gpt3, gpt4, llama} have demonstrated exceptional versatility and proficiency in modeling text content in various domains and tasks. In everyday applications, ChatGPT~\cite{chatgpt} and GPT-4~\cite{gpt4} are significantly well-performed for general purposes such as language translation, question answering, and document composition. In addition, they excel in handling extensive long-range contexts. For example, GPT-4 can process a maximum input of 32,000 tokens. More recently, Xie \etal~\cite{visorgpt} demonstrated that visual priors, such as object locations, shapes, and their relations, can be effectively modeled through generative pre-training. They represent images using a set of visual locations, such as object bounding boxes, human keypoints, and semantic masks. These visual elements are simply discretized and then tokenized, same as text, into a series of tokens, which can be learned through generative pre-training. In light of this, our goal is to investigate the feasibility of learning the long-form video prior, focusing on aspects of people, objects, and their interactions over time, through generative pre-training.

Instead of directly learning the long-form video prior in pixel space, we alternatively model it in the token space by maximizing the predicted probability of the next tokens in a sequence. Compared to pure image pixels, object bounding boxes and whole body keypoints sufficiently represent the key information in long-form videos. Consider long-form movies as an example, the bounding boxes can indicate the blocking of characters and the whole body keypoints can indicate their facial expressions and actions. Hence, it becomes possible to model the long-form video prior from these visual locations. Specifically, we propose to represent each movie as a sequence of tokens, in which continuous visual locations are discretized as words and then tokenized with the textual descriptions. In this way, a transformer decoder can be employed to learn the long-form video prior from these tokens through generative pre-training. In addition, instruction tuning can be integrated into the training process, after which novel videos can be sampled from the learned prior according to the user-provided instructions.

Due to the scarcity of suitable data, we contribute a new movie dataset namely \textbf{Storyboard20K} to facilitate the learning of long-form video prior. As movies represent an artistic expression of human stories, they typically encompass a wide range of concepts such as people, objects, and their interactions over time, thereby capturing the distinctive characteristics and challenges inherent in movies. Specifically, according to previous studies~\cite{sc2st, tevis}, we collect 20,310 movie storyboards from MovieNet~\cite{movienet} and LSMDC~\cite{lsmdc} in total and each currently contains only multiple shots (keyframes) associated with corresponding textual descriptions. To facilitate the learning of long-form video prior via generative pre-training, several elements are taken into consideration such as characters, blocking, actions, and camera transitions. Correspondingly, we annotate the bounding boxes, whole body keypoints, and consistent ID of each character across each shot. It is first automatically processed by current algorithms~\cite{CascadeRcnn, vipnas} and further calibrated by human annotators. As shown in Fig.~\ref{fig:teaser} (a), across various shots, the same character has a consistent and correct ID, and their blocking and actions are represented by bounding boxes and whole body keypoints. Note that, apart from characters, the bounding boxes of film sets such as cars and buildings are also annotated, which comes from nouns in textual descriptions or scene associations. In addition, we provide each storyboard with more summative information such as the title, genre, expression, scene, etc. Annotated samples are provided in Fig.~\ref{fig:dataset}. Experimental results on the proposed dataset demonstrate that our approach has great potential to learn the long-form video prior via generative pre-training.

The main contributions of this paper are summarized as:
\begin{itemize}
    \item We view extensive concepts in long-form videos as an implicit prior and propose to learn the prior via generative pre-training.

    \item Instead of operating on pixel space, we propose to represent long-form videos as sequences of tokens, in which long-form video prior can be effectively learned via generative pre-training. 

    \item We present a new dataset, \emph{i.e.,} \textbf{Storyboard20K}, to serve as a representative to facilitate the learning of GPT on long-form videos. Experimental results on the dataset validate that our approach has great potential to model the prior for vision generation tasks.
    
\end{itemize}

\section{Related Works}
\label{sec:relate_works}
\subsection{Video Generation}
In recent years, advanced progress in diffusion models has motivated a wave of research in video generation, encompassing diverse areas such as unconditional video generation~\cite{vdm, latentvdm, videofusion}, text-to-video generation~\cite{video1, make-a-video, animatediff, show1}, and text-guided video editing~\cite{tune-a-video, structure, editavideo, videofusion}. Notably, make-a-video~\cite{make-a-video} pioneered the direct extension of sophisticated text-to-image techniques to text-to-video, leveraging extensive knowledge without the need for paired text-video data. However, these approaches mainly focus on short video generation and still encounter challenges such as temporal consistency. Hence, this line of methods may still not be suited for long-form scenarios due to increased complexity in human interactions and view changes.

\subsection{Learning from Movies}
In an era marked by the rapid evolution of media technologies, including television and film, the global popularity of cinematic experiences is more evident than ever. Over the past few years, there has been a growing focus on research related to learning from movies~\cite{lsmdc, autoad, mad, tevis, sc2st}. The long-time coverage and complicated events involved in movies make it difficult for neural networks to comprehensively understand and generate. To facilitate learning from movies, there are many movie datasets have been proposed. For instance, Rohrbach \etal create the LSMDC~\cite{lsmdc} dataset, which includes nearly 128K video clips with transcribed audio description (AD) sourced from 200 movies. Compared to natural captions of images, AD provides a dense narrative of important events that happened in the movie, including key information about where, who, and what. More recently, a holistic dataset, \emph{i.e.,} MovieNet~\cite{movienet}, has been proposed for movie understanding. It contains 1,100 movies with a large amount of multi-modal data, \emph{e.g.,} trailers, and plot descriptions. In the past few years, many works have been proposed for movie understanding or generation. Han \etal~\cite{autoad} proposed an Audio Description (AD) model to automatically generate AD for current movie clips. Tian \etal~\cite{sc2st} and Gu \etal~\cite{tevis} proposed to generate movie storyboards by retrieving relevant visual frames from a pre-defined database. Based on the aforementioned studies, we created the Storyboard20K with more fine-grained annotations to facilitate the learning of long-form video prior.

\subsection{Language Modeling}
Language modeling, a crucial task in natural language processing, involves estimating the probability of a specific sequence of words occurring in a sentence. Over the past decade, numerous unsupervised language modeling approaches have been proposed, without the requirement of manual labels. Devlin \etal~\cite{bert} pioneered the pre-training of bidirectional transformers, introducing BERT (Bidirectional Encoder Representations from Transformers) through masked language modeling. As a pre-training approach, BERT has showcased remarkable proficiency across diverse downstream tasks such as translation and question-answering. Following this approach, subsequent models like Albert~\cite{albert} and Roberta~\cite{roberta} have extended the capabilities of bidirectional transformers. Different from this line of methods, Radford \etal~\cite{gpt1} proposed the generative pre-training (GPT), which employs the transformer decoder to sequentially model the probability of the next tokens by maximizing the likelihood of given sentences. The GPT series, including GPT-2~\cite{gpt2}, GPT-3~\cite{gpt3}, and the latest GPT-4~\cite{gpt4}, has been successively proposed, demonstrating unparalleled language modeling abilities. With large-scale data and an increased number of model parameters, the GPT series has proven its remarkable capacities in language modeling, excelling in various natural language processing tasks. Employing generative pre-training, this work endeavors to demonstrate the feasibility of learning long-form video prior from the proposed Storyboard20K dataset.

\section{Storyboard20K}
To facilitate the modeling of long-form video prior through generative pre-training, we propose to represent long-form videos from three main aspects, \emph{i.e.,} \textbf{textual descriptions}, \textbf{character-centric} and \textbf{film-set-centric visual locations over time}. This structured representation of key information in long-form videos can be potentially consumed by generative pre-training. Due to the scarcity of suitable data, we contribute a new dataset Storyboard20K curated from movies, encapsulating synopses, character-centric, film-set-centric, and summative annotations. In the following, we will provide an overview of data collection and annotations. Subsequently, we will present an overall data analysis and discuss the potential properties.

\begin{table}[t]
    \centering
    \setlength\tabcolsep{8pt}
    \resizebox{\linewidth}{!}{
    \begin{tabular}{cccccccc}
        \toprule
         \multirow{2}{*}{\#Storyboards} & \multirow{2}{*}{\#Shots} & \multicolumn{2}{c}{\#Bboxes per Storyboard} & \multirow{2}{*}{Avg. Synopsis Length} & \multirow{2}{*}{\textbf{Avg. Duration}}\\
        \cmidrule(lr){3-4} 
         &  & Characters & Film sets &  \\
         \midrule
         20,310 & 149,528 & $\sim$15.7 & $\sim$8.9 & $\sim$381.4 words & $\sim$\textbf{33.9 seconds} \\
         
         \bottomrule
    \end{tabular}}
    \caption{An overview of the proposed Storyboard20K. It includes $\sim$20K storyboards sourced from MovieNet~\cite{movienet} and LSMDC~\cite{lsmdc} with $\sim$150K shots (key frames) in total. For each storyboard, there are around 13.8 and 4.7 annotated bounding boxes for characters and film sets, respectively. The average duration of these movie storyboards is around 33.9 seconds, which ensures long-time coverage and large view variation.}
    \label{tab:dataset}
\end{table}

\begin{figure*}[t]
    \centering
    \includegraphics[width=\linewidth]{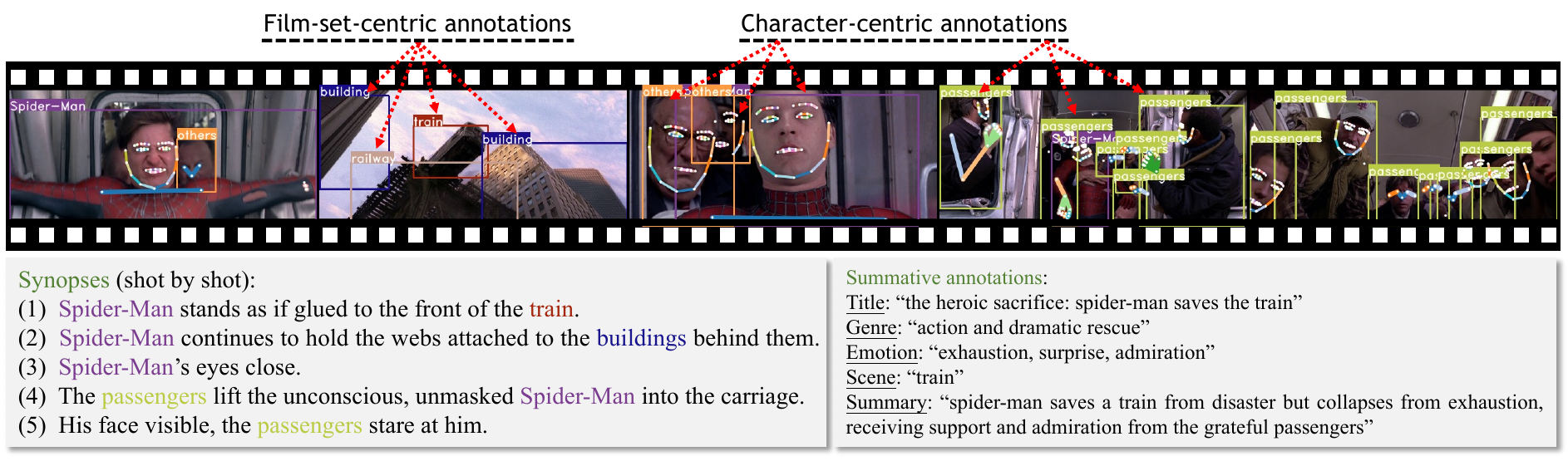}
    \caption{Annotated samples (part of a storyboard) of the proposed Storyboard20K. Our dataset involves three main annotations, \emph{i.e.,} (i) character-centric (whole body keypoints and bounding boxes with consistent IDs), (ii) film-set-centric (bounding boxes), and (iii) summative (texts) annotations. It also includes condensed (as illustrated in Fig.~\ref{fig:teaser}) or shot-by-shot descriptions.}
    \label{fig:dataset}
\end{figure*}

\textbf{Data Collection}. According to previous studies~\cite{tevis, sc2st}, we collect 20,310 movie storyboards with $\sim$150K shots (keyframes) sourced from MovieNet~\cite{movienet} and LSMDC~\cite{lsmdc} to form the Storyboard20K dataset. During the formation of these storyboards, each candidate movie clip was condensed into multiple representative shots (frames) based on the textual descriptions, usually within a range of 3 to 10 shots per storyboard. As we know, in filmmaking, all shots in a storyboard generally share the same storyline. Hence, when constructing a storyboard, a strong emphasis was placed on ensuring the story coherence of each shot by algorithms~\cite{clip} and manual screening, and outliers that deviated from the intended storyline were systematically removed. The movie clips typically have an average duration of 33.9 seconds, resulting in \textit{long-form time coverage and large view variation} for each storyboard. We present a part of a movie storyboard in Fig.~\ref{fig:dataset}. One can observe that multiple shots of the storyboard share the same storyline, \emph{i.e.,}, ``Spider-Man saves the train and passengers'' and the transitions across various shots exhibit large variations. Apart from visual shots, the dataset includes two types of synopses: `condensed description' and `shot-by-shot description'. Illustrative examples are provided in Figs.~\ref{fig:teaser} and ~\ref{fig:dataset}. Notably, the condensed descriptions provide a concise narration of the storyboard's main contents (Fig.~\ref{fig:teaser}) and the shot-by-shot descriptions offer more detailed information (Fig.~\ref{fig:dataset}), both probably encompassing main characters, actions and facial expressions. 

\textbf{(i) Character-Centric Annotation}. While filming a storyboard, the director should thoroughly and carefully draft the blocking, actions, and facial expressions of primary and extra roles. The aforementioned elements with artistic shot transitions across various shots significantly determine the quality of final visual effects. As only textual descriptions cannot explicitly and completely demonstrate the above elements, we provide fine-grained annotations of blocking, actions, and facial expressions of characters in spatial by annotating object-bounding boxes and whole body keypoints. Besides, beyond a single shot, as the transitions of blocking across various shots are very important for the quality of presentation, the IDs of characters are also consistently annotated to remain the same throughout the time. To obtain such annotations in high quality and efficiently, we design a two-stage approach with \textbf{automatic tagging and manual calibration}. First, we employ state-of-the-art object detection methods~\cite{CascadeRcnn, vipnas} to detect human bounding boxes and whole body keypoints of each shot. For each actor in a movie, we collect a set of headshots according to the cast list presented in IMDB. For a storyboard, face detection~\cite{mtcnn} and identification~\cite{facenet} methods are used to tag the same actor with the same IDs across shots, and the IDs are sourced from the current shot descriptions. However, many artistic effects are involved in movie frames, such as darkness and low-key lighting, and the presence of numerous background and close-up shots of characters, which potentially render the current algorithms ineffective. To ensure the annotations are of high quality, we ask annotators to manually calibrate the character IDs across shots at the storyboard level. We ensure that the annotators possess a strong understanding of English, and they are tasked with manually correcting misaligned bounding boxes and rectifying incorrect IDs on a per-storyboard basis. We provide some samples in Figs.~\ref{fig:teaser} and \ref{fig:dataset}. It can be observed that the main character ``Spider-Man'' is consistently annotated with the same ID, bounding boxes, and whole body keypoints across various shots. In addition, extra roles such as the passengers are also precisely annotated.

\textbf{(ii) Film-Set-Centric Annotation}. Apart from human-centric interaction, determining film sets such as props, background, and structures in the current storyboard is also significant in affecting the final presentation of a movie. For this reason, we have unrestrictedly annotated the bounding boxes of film sets that are related to the current storyboard from the perspective of shot descriptions or visual frames. Imagine you are shooting a storyboard about ``two middle-aged men having drinks at a bar''. A professional director would begin with an extreme close-up shot of the interior of the bar, possibly including a resident singer and their band performing. Then the shot transitions to the main characters, with a cocktail glass nearby, before moving on to the rest of the story. In this example, the resident singer, band performers, musical instruments, and the cocktail glass are the film set, and annotators are tasked with including these objects with names and bounding boxes in the annotations. Similar examples are illustrated in Fig.~\ref{fig:dataset}. One can observe that film sets exist in the scripts such as ``buildings'' and ``train'' in the second shot are accordingly annotated. Besides, the ``railway'' which is not described in the scripts but related to the ``train'' is also annotated. This kind of annotation would provide more sufficient association among various categories and scenes.

\textbf{(iii) Summative Annotation}. Beyond the narrative descriptions of storyboards, we provide more summative annotations. Specifically, for each movie storyboard, we conclusively give the ``title'', ``genre'', ``emotion'', ``scene'', and a rewritten short ``summary''. We showcase an example on the bottom-right of Fig.~\ref{fig:dataset}. This kind of summative annotations can be considered as synopses drafted by a movie director, which can be further used for classification or as human instructions for fine-tuning.

\begin{figure}[t]
    \centering
    \includegraphics[width=\linewidth]{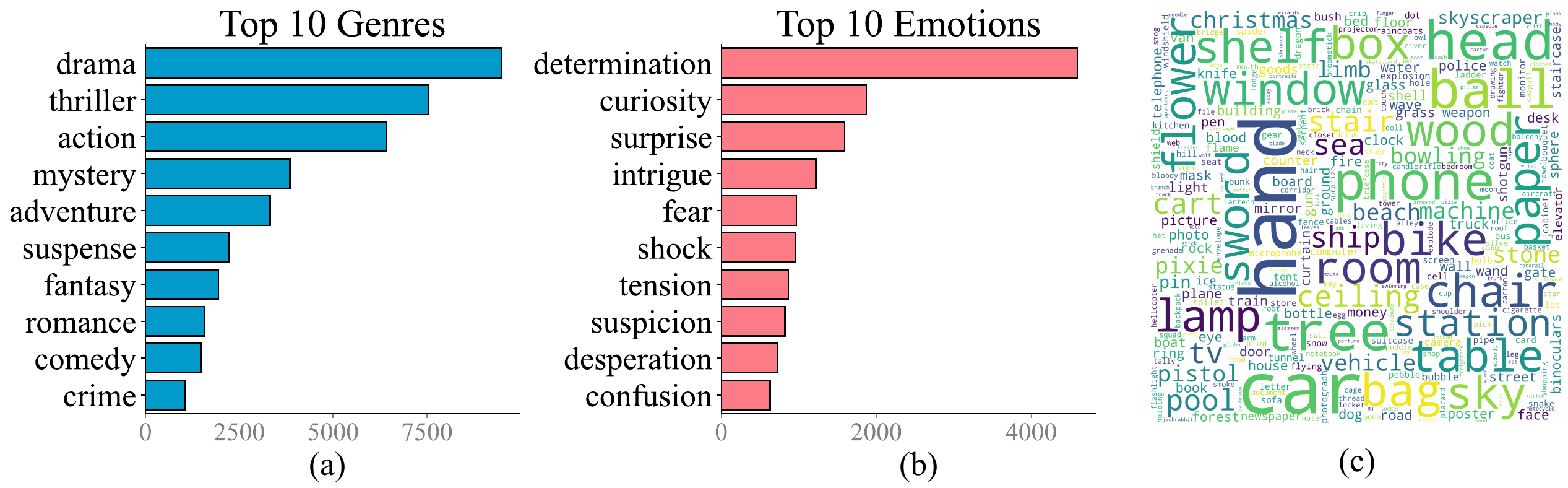}
    \caption{Summative analysis. (a) and (b) are the statistics of the top 10 genres and emotions. (c) is the visualization of the most frequent film sets (a subset of 300 categories) using a word cloud.}
    \label{fig:summative_analysis}
\end{figure}

\textbf{Dataset Analysis}. An overview of Storyboard20K is presented in Table~\ref{tab:dataset}. There are 20,310 movie storyboards with around $\sim$150K movie frames sourced from around 400 movies. For the assessment of story coherency and textual diversity, previous studies \cite{movienet, lsmdc, sc2st, tevis} provide a comprehensive analysis to demonstrate the superiority of the source data. Beyond this aspect, we will analyze the proposed Storyboard20K from the above three movie-centric annotations. First, we present the statistics of top-10 coverage of genres, and emotions in Figs.~\ref{fig:summative_analysis} (a) and (b). One can observe that the summative annotations cover most common genres such as drama, action, and comedy, and emotions such as surprise, fear, and confusion. Table~\ref{tab:dataset} shows the statistics of character and film set annotations. As observed, around 500K bounding boxes ($\sim$24.6 bounding boxes average per storyboard) are annotated including 319K for characters and 180K belonging to film sets. From our statistics, there are a total of around 6K different categories of film sets (synonyms may be included), which vary from indoor to outdoor and things to stuff. We give a visualization of 300 categories using word clouds in Fig.~\ref{fig:summative_analysis} (c). This validates that film set annotations in the proposed Storyboard20K are diverse, and inclusively cover the common objects in everyday life.

\begin{figure*}[t]
    \centering
    \includegraphics[width=\linewidth]{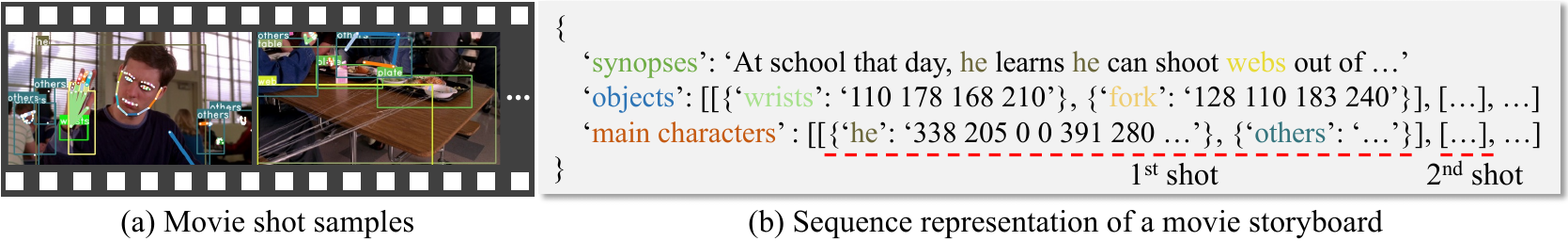}
    \vspace{-10pt}
    \caption{Illustration of the prompt designs to represent a movie storyboard consisting of multiple shots.}
    \label{fig:prompt}
    \vspace{-10pt}
\end{figure*}

\textbf{Discussion}. In this paper, the proposed Storyboard20K aims to provide high-quality and professional movie storyboard data and annotations to facilitate the research on learning long-form video prior. Beyond that, as large-scale textual descriptions with corresponding bounding boxes, whole body keypoints, and consistent IDs across various shots, the proposed dataset can be also employed to benefit AI-assisted filmmaking, visual grounding, character recognition and re-identification, and tracking. Further, the sufficient multi-modal information in our dataset can be a kind of source for training large multi-modal models for both understanding and generation, which can significantly enhance their capacities toward the movie domain.

\textbf{Publicity and Copyright}. Both source movie clips/frames of LSMDC and MovieNet can be downloaded from their official websites\footnote{https://sites.google.com/site/describingmovies/download?authuser=0}\textsuperscript{,}\footnote{https://opendatalab.com/OpenDataLab/MovieNet/tree/main/raw}. We will only release the annotations to avoid any potential concerns related to movie copyright issues of the proposed Storyboard20K.

\section{Method}
\label{sec:method}
In this section, we first formulate the problem of learning long-form video prior ($\S$~\ref{sec:pf}). Following that, we introduce our prompt designs ($\S$~\ref{sec:pd}) to encode key information of videos such as textual descriptions and visual locations into token space. Subsequently, we present the model architecture and learning objective ($\S$~\ref{sec:po}) to learn the long-form video prior. In the end, we illustrate how to sample novel outputs from the learned long-form video prior ($\S$~\ref{sec:sample}). Note that we employ movie storyboards as representative of long-form videos.

\subsection{Problem Definitions}
\label{sec:pf}

This work focuses on learning long-form video prior in the token space, in which pixel information in movie storyboards is not considered. When object bounding boxes and whole body human keypoints are discretized and then tokenized the same as texts, these tokens $x$ can be assumed to follow a joint probabilistic prior distribution $p(x)$. As many hidden variables are jointly involved in $p(x)$, it is often inaccessible in practice. This work proposes to approximate it by learning a neural network $q_\theta$ via likelihood maximization of each movie storyboard. In this way, novel outputs $\tilde{x}$ can be sampled from learned $q_\theta(x)$.

\subsection{Prompt Designs for Long-form Videos} 
\label{sec:pd}

In the area of natural language processing, various tasks such as translation, document composition, and question-answering share the same 
carrier, \emph{i.e.,} text, which can be operated on the same token space. Unlike pure language, long-form videos such as movie storyboards involve extensive multi-modal information including textual descriptions and continuous visual locations of characters and film sets such as bounding boxes and keypoints. One of the characteristics of these kinds of visual locations is that they are highly structured. For example, a bounding box is usually represented by coordinates of top-left and bottom-right corners and a human action can be generally represented using 133 keypoints with x-axis and y-axis coordinates~\cite{wholebodykeypoint}. As illustrated in previous study~\cite{visorgpt}, continuous numbers can be quantized into $m$ bins such that each visual location can also be naturally represented as a set of discrete words and only $m$ integers are required to be accordingly added to the standard vocabulary. By doing so, each number can be represented by a word in the range of $[1,m]$ and each movie storyboard can be correspondingly represented by a sequence of words.

Recent LLMs such as ChatGPT~\cite{chatgpt} and GPT-4~\cite{gpt4} have demonstrated their proficiency in giving structural responses. Here, to facilitate the learning and decoding process, we propose the structural prompt designs to encode the movie storyboard. Specifically, as shown in Fig.~\ref{fig:prompt}, a movie storyboard $x$ involving $N$ shots can be processed into a sequence $s=\textsc{Prompt}\left(x\right)$, which is structured in JSON format. Specifically, it includes three main elements, \emph{i.e.,} `synopses', `objects', and `main characters'. For `objects', there are $N$ sub-lists representing various shots, which consists of the discrete location of film sets such as \{`fork': [128 110 183 240]\}. For `main characters', it involves $N$ sub-lists with discrete whole body keypoint information such as \{`he': [338 205 0 0 391 280 ...]\}, in which every two integers represents a keypoint and the order follows the previous study~\cite{wholebodykeypoint}. More specifically, to reduce the consumption of tokens and make an emphasis on the main characters, whole body keypoint representation is only for the characters with an area larger than $96^2$ pixels, otherwise, we use 17 keypoints or bounding boxes to represent the characters. As shown in Fig.~\ref{fig:prompt}, the character `he' is represented by whole body keypoints. Note that the start and end of tokens are correspondingly added to the sequence.

Limited by the computational resources and efficiency, the maximum tokens of the input sequence are usually not any length. In this work, we set the number of maximum input tokens as 2,560 to involve more shots of a storyboard. 


\subsection{Modeling Prior via Generative Pre-Training}
\label{sec:po}

\quad \textbf{Model Architecture}. Large language models such as GPT-2~\cite{gpt2}, GPT-3~\cite{gpt3}, ChatGPT~\cite{chatgpt}, and VisorGPT~\cite{visorgpt} have demonstrated their remarkable capacities in modeling any kind of text content such as books and visual locations. Following their model designs, we employ the transformer decoder as our model to learn the long-form video prior. Specifically, we adopt the open-source GPT-2 (base).

\textbf{Learning Objective}. Each movie storyboard $x$ will be processed into a textual sequence $s$ then each sequence is tokenized into a set of $n$ tokens $u = \{u_1, u_2, \cdots, u_n\}$. In this way, we employ the standard language modeling objective in ~\cite{gpt2, visorgpt} to learn the long-form video prior by maximizing the following likelihood of each sequence:
\begin{equation}
    \mathcal{L} = \sum_{i} \text{log} p(u_i | u_{i-k}, \cdots, u_{i-1}; \Theta), 
\end{equation} 
where the context window size is denoted as $k$, and $p(\cdot | \cdot)$ represents the conditional probability  modeled by the neural network $\Theta$. We adopt stochastic gradient descent to train the neural network. 

\subsection{Sampling from long-form video prior}
\label{sec:sample}

Once the long-form video prior is learned, novel outputs $\tilde{x}\sim q_\theta$ can be accordingly sampled. Here, we provide some examples of sampling: i) \textit{Unconditional sampling}. Given only the start of token, we can randomly sample novel storyboards from the learned prior $q_\theta$. ii) \textit{Conditional sampling on the synopsis}. Given the synopsis such as ``a group of people are talking together in the room'', we can correspondingly sample a movie storyboard relevant to the textual description. iii) \textit{Conditional sampling on the instruction}. Beyond purely learning on the tokenized movie storyboard, the instruction-storyboard pairs such as ``Could you please develop a movie storyboard that takes place in a room?'' based on the summative annotations are generated to train the model. This ensures a more free-form sampling using instructions from the learned prior.

\section{Experiments}
\label{sec:experiments}

\subsection{Experimental Setup}
\quad \textbf{Dataset}. We utilize the proposed Storyboard20K dataset for learning the long-form video prior. Specifically, the storyboard dataset is divided into three subsets: \textit{train}, \textit{testA}, and \textit{testB}, which comprise 18,207, 905, and 1,198 storyboards, respectively. The distinction between \textit{testA} and \textit{testB} is that \textit{testA} consists of condensed descriptions only (as shown in Fig.~\ref{fig:teaser}), while \textit{testB} includes shot-by-shot descriptions exclusively (as shown in Fig.~\ref{fig:dataset}).

\textbf{Evaluation Metrics}. In this work, movie storyboards serve as representatives of movies for learning the prior, encompassing multi-modal information in the form of textual descriptions and visual locations. To comprehensively assess the proposed approach, we employ classic NLP metrics, including Rouge-L~\cite{rouge}, BertScore~\cite{bertscore}, and Perplexity, to evaluate the generated sequences of movie storyboards. Additionally, FID~\cite{fid} is used to evaluate the generated movie storyboards from the perspective of visual locations. Specifically, we adopt layouts of MSCOCO~\cite{coco} to train a classifier that can be used to extract features for computing FID between two sets of layouts.

\textbf{Implementation Details}. Following~\cite{visorgpt}, we set the number of bins $m$ as 512, indicating that all coordinates of bounding boxes and human keypoints will be re-scale to a resolution of $512\times512$ and then discretized as integers. In this way, 512 integers are accordingly added to the standard vocabulary as independent words. We set the number of maximum input tokens as 2,560. The maximum number of shots is set as 4 when the sequence includes keypoints, otherwise, it is set as 10. Instead of using all 133 whole-body body keypoints, we only sample 93 keypoints by eliminating some facial and hand points (details are provided in supplementary materials) to form the sequences. During training, we employ the AdamW optimizer with a maximum learning rate of 5e-5, which will decay to 5e-6 linearly. Total training iterations are set as 80,000. Besides, DeepSpeed strategy with 16-mixed precision is adopted to reduce the memory cost. All experiments are conducted on 8 Tesla V100 with a memory of 32GB. More details are presented in supplementary materials.

\begin{table}[t]
    \centering
    \resizebox{\linewidth}{!}{
    \begin{tabular}{lc|c|c|c|c}
        \toprule
        Methods & OpenWebText & Storyboard20K & Rouge-L $\uparrow$ & BertScore $\uparrow$ & Perplexity $\downarrow$ \\
         \midrule
         GPT-3.5-turbo & - &  - & 20.4 / 17.6 & 96.8 / 95.7  & - \\
         \rowcolor{mygray}
         Ours (GPT-2) & $\times$ & \checkmark & 19.3 / 23.7 & 96.7 / 96.1 & 1.2 / 1.5 \\
         \rowcolor{mygray}
         Ours (GPT-2) & \checkmark & \checkmark & 19.6 / 26.7 &  97.1 / 96.3 & 1.1 / 1.3 \\
         \bottomrule
    \end{tabular}}
    \caption{Quantitative comparison with GPT-3.5-turbo using textual metrics such as Rouge-L, BertScore, and Perplexity. To assess GPT-3.5's performance, we employ in-context learning examples to enable it to generate movie storyboards comprising textual descriptions and visual locations. The evaluation is conducted on \textit{testA} and \textit{testB}, with their performance presented in the format of \textit{testA} / \textit{testB} for clarity.}
    \label{tab:textual_eval}
    \vspace{-20pt}
\end{table}

\begin{table}[t]
    \centering
    \resizebox{\linewidth}{!}{
    \begin{tabular}{lc|c|c|cc}
        \toprule
         Methods & OpenWebText & VisorGPT dataset~\cite{visorgpt} & Storyboard20K & FID on \textit{testA} $\downarrow$ & FID on \textit{testB} $\downarrow$ \\
         \midrule
         GPT-3.5-turbo & - & - & - & 23.08 & 35.14 \\
         \midrule
         Ours (GPT-2) & \checkmark & $\times$ & \checkmark & 5.43 & 29.16 \\
         Ours (GPT-2) & $\times$ & $\times$ & \checkmark & 4.42 & 27.09 \\
         \midrule
         \rowcolor{mygray}
         Ours (GPT-2) & $\times$ & \checkmark & \checkmark & \textbf{2.94} & \textbf{23.40} \\
         \bottomrule
    \end{tabular}}
    \caption{Quantitative comparison with GPT-3.5-turbo
    based on FID. Specifically, we provide in-context learning examples for GPT-3.5-turbo to generate movie storyboards with textual descriptions and visual locations. }
    \label{tab:visual_eval}
    \vspace{-15pt}
\end{table}

\subsection{Quantitative Results}
We present textual evaluation results in Table~\ref{tab:textual_eval}, utilizing metrics such as Rouge-L, BertScore, and Perplexity. To establish a benchmark, we compare our models against advanced Large Language Models (LLMs) like GPT-3.5. Pre-training on extensive web-scale datasets, GPT-3.5 possesses remarkable zero-shot capabilities in many tasks such as translation and question answering. In this comparative analysis, we provide in-context learning examples for GPT-3.5, tasking them with generating movie storyboards with diverse characters, film sets, and shot variations. Due to the unavailability of GPT-3.5 model, they cannot be fine-tuned on our proposed Storyboard20K dataset. One can observe in the table that our models outperform GPT-3.5 across textual metrics of Rouge-L, and BertScore. This validates our method's ability to effectively model the long-form video prior, resulting in the generation of coherent and contextually appropriate movie storyboards.

Besides, we provide a visual evaluation in Table~\ref{tab:visual_eval} using the FID metric. This metric examines whether the generated layouts such as bounding boxes and keypoints are similar to the real ones, exhibiting how well can the proposed method model the long-form video prior. One can observe from the table that GPT-3.5 can achieve FIDs of 23.08 and 35.14 on \textit{testA} and \textit{testB} sets, respectively, through in-context learning. In contrast, our models outperform them by a large margin on both \textit{testA} and \textit{testB} sets. An interesting phenomenon can be found in that with pre-training on the large-scale dataset used in VisorGPT, our GPT-2 gains significant improvement over the model-only training on Storyboard20K.

\begin{wraptable}{l}{7cm} 
\vspace{-20pt}
\setlength\tabcolsep{5pt}
\resizebox{\linewidth}{!}{%
\begin{tabular}{lccc}
        \toprule
        Methods & \#Sampling & \#Valid Sampling & Success Rate (\%) \\
         \midrule
         GPT-3.5-turbo & 1,000 & 535 & 53.5 \\
         \rowcolor{mygray}
         Ours (GPT-2) & 1,000 & 925 & 92.5 \\
         \bottomrule
    \end{tabular}}
\vspace{-10pt}
\caption{Decoding success rate. 1,000 samplings are conducted to examine the success rate of decoding into movie storyboards.}
    \label{tab:decoding_success_rate}
\vspace{-16pt}
\label{tab:impact_theta_alpha}
\end{wraptable}
As introduced in Sec.~\ref{sec:pd}, we encode each movie storyboard into JSON format for convenient decoding. A question arises: can the proposed method effectively learn and infer outputs in this format, ultimately being successfully decoded into movie storyboards? As shown in Table~\ref{tab:decoding_success_rate}, where we compare the decoding success rates with GPT-3.5. In particular, we conducted 1,000 samplings and counted the number of valid sequences successfully decoded into movie storyboards. The table shows that our model achieves a relatively high decoding success rate, \emph{i.e.,} 92.5\%, compared to GPT-3.5, \emph{i.e.,} 53.5\%. This result suggests that the proposed prompt designs can be effectively learned, facilitating convenient decoding.

\begin{figure*}[t]
    \centering
    \includegraphics[width=\linewidth]{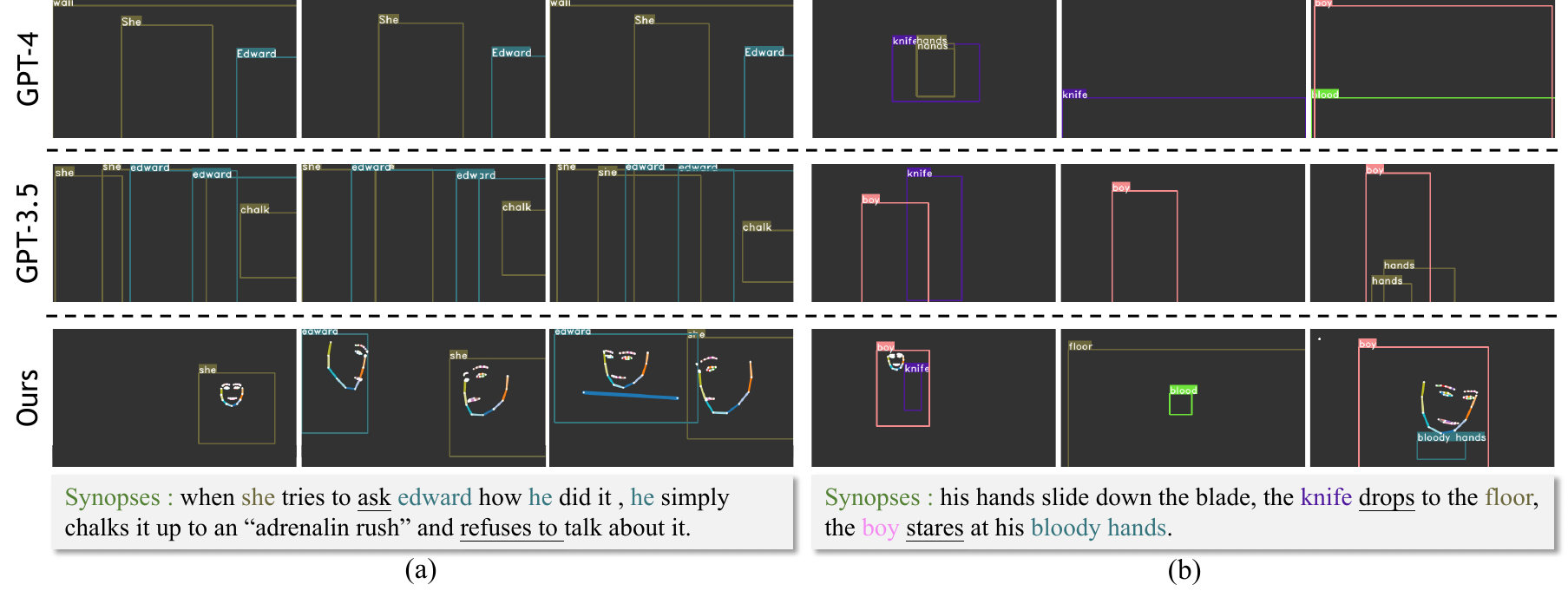}
    \vspace{-20pt}
    \caption{Visualization of movie storyboards sampled from the long-form video prior learned by our approach and predicted by GPT-4 and GPT-3.5. We provide in-context learning examples for GPT-3.5/4, \textbf{tasking them with generating movie storyboards with diverse characters, film sets, and shot variations}. One can observe that the movie storyboards generated by our model exhibit large view changes across various shots while maintaining a strong alignment with the provided scripts. For instance, given the script ``his hands slide down the blade, the knife drops to the floor, the boy stares at his bloody hands'', the generated movie storyboard initially presents a middle shot of the boy holding a knife. Subsequently, the second shot captures the blood on the floor, and in the final shot, the focus shifts to the boy staring at his bloodied hands.}
    \label{fig:demo}
    \vspace{-10pt}
\end{figure*}

\begin{wrapfigure}{l}{7cm} 
\vspace{-18pt}
\includegraphics[width=\linewidth]{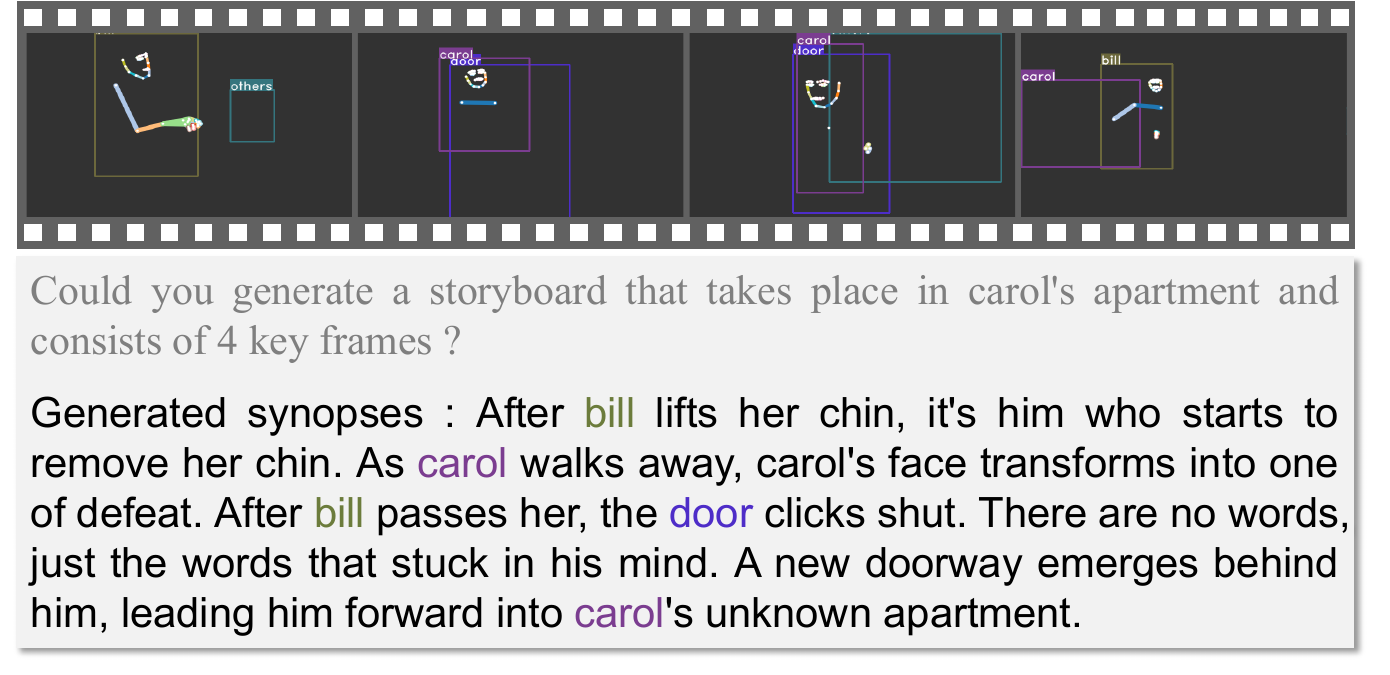}
\vspace{-20pt}
\caption{Visualization of movie storyboards sampled from the long-form video prior learned by our approach following the instruction of featuring ``Peter'' as the main character.}
    \label{fig:demo2}
\vspace{-14pt}
\end{wrapfigure}


\subsection{Visualization Results}
In addition, we also present a visualization of movie storyboards sampled from the learned long-form video prior in Fig.~\ref{fig:demo}. One can observe that our model can generate movie storyboards that closely align with the scripts. In addition, our generated movie storyboards exhibit large view changes over time according to the scripts. One can observe in Fig.~\ref{fig:demo} (a), there are two primary roles, \emph{i.e.,} `she' and `Edward', in the script and the generated movie storyboard can capture the essence of the plot and demonstrate a good alignment. Specifically, the first shot depicts the appearing of `she', setting the stage for the narrative. Subsequently, in the second shot, our model accurately interprets the script's instruction for `she' to `ask', predicting a set of keypoints that position `she' facing the other role `Edward', showing the action of asking for information. Similarly, a similar phenomenon can be observed in the conditional sampling depicted in Fig.~\ref{fig:demo} (b). Given the script ``his hands slide down the blade, the knife drops to the floor, the boy stares at his bloody hands'', the generated movie storyboard follows the script's narrative. The initial shot focuses on a middle shot of the boy holding a knife, followed by a shot capturing the blood on the floor, and ending with a final shot of the boy staring at his bloody hands. Fig.~\ref{fig:demo2} also presents a sampled storyboard following the instruction. These visual results demonstrate the capability of the proposed approach to learn intrinsic priors within movies and generate coherent storyboards with the script. 

\begin{wrapfigure}{l}{7cm} 
\vspace{-14pt}
\includegraphics[width=\linewidth]{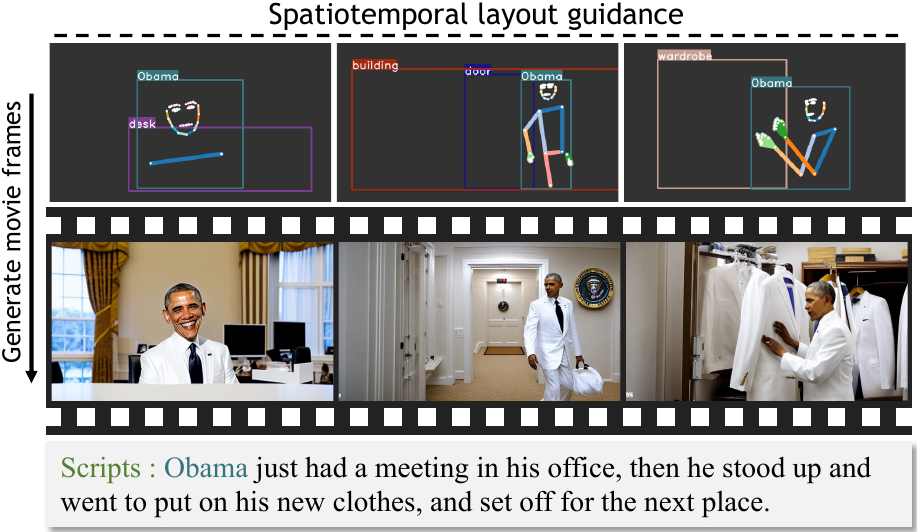}
    \vspace{-20pt}
    \caption{Pixel frames synthesized from a storyboard.}
    \vspace{-18pt}
    \label{fig:demo3}
\end{wrapfigure}
Visual comparisons among GPT-4, GPT-3.5, and our model are presented in Fig.~\ref{fig:demo}. Note that we prompt GPT4/3.5 to generate diverse storyboards with large variations. Given a script ``when she tries to ask Edward how he did it, he simply chalks it up to an “adrenalin rush” and refuses to talk about it.'', both GPT-4 and GPT-3.5 infer the main role `she' and `Edward' and some related film sets. However, the generated three shots are extremely similar, exhibiting no variations or view changes requested by our prompts. When provided with a script ``his hands slide down the blade, the knife drops to the floor, the boy stares at his bloody hands'', both GPT-4 and GPT-3.5 can comprehend the main role ``boy'' and related film sets such as a knife. However, the generated storyboard does not exhibit sufficient coherence, lacking alignment with the given script. In contrast, movie storyboards sampled from our model demonstrate accurate reasoning of the main role and film sets. Moreover, the generated storyboards align closely with the script, unfolding in a logical sequence shot by shot.

\subsection{Applications}
The storyboard can serve as spatiotemporal guidance for controllable image/video diffusion models such as ControlNet~\cite{controlnet} and UniControl~\cite{unicontrol}, enabling the generation of frames that maintain consistency in characters, objects, and their interactions. We illustrate an example in Fig.~\ref{fig:demo3}. Additionally, since our model is trained on movies, the generated storyboard can be a valuable resource for amateurs, providing guidance on blocking and assisting them in the creation of their own movies.

\section{Conclusion}
\label{sec:conclusions}

In this study, we investigated the feasibility of utilizing generative pre-training (GPT) to learn the long-form video prior. Instead of directly operating on pixel space, we adopted visual locations such as bounding boxes and keypoints to represent key information in movies. To facilitate this exploration, we contributed a new movie dataset \textbf{Storyboard20K} consisting of scripts, shot-by-shot keyframes, and fine-grained annotations of film sets and characters with consistent IDs, bounding boxes, and whole-body keypoints. By discretizing and tokenizing these visual elements and the script into token space, we demonstrate that long-form video prior can be successfully learned by generative pre-training --- maximizing the likelihood of each sequence. 

%
%
\bibliographystyle{splncs04}
\bibliography{main}

\clearpage
\setcounter{page}{1}
\setcounter{section}{0}
\setcounter{figure}{0}
\setcounter{equation}{0}

\section{Additional Data Samples}

\begin{figure}[h]
    \centering
    \includegraphics[width=1\linewidth]{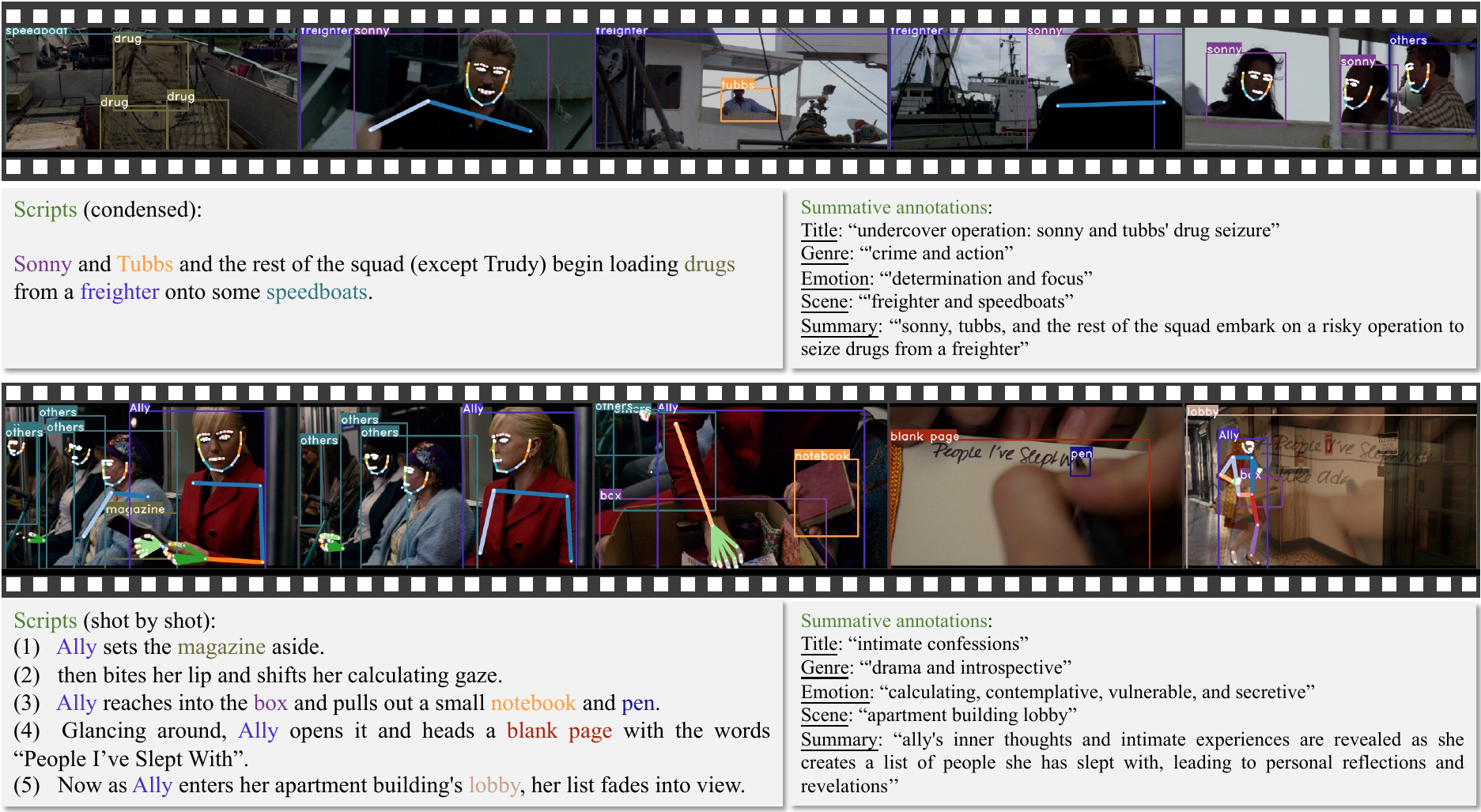}
    \caption{Additional storyboard samples.}
    \label{fig:supp_dataset}
\end{figure}

We provide additional samples of the proposed Storyboard20K in Fig.\ref{fig:supp_dataset}. Each movie storyboard in the dataset exhibits extensive temporal coverage, resulting in significant view changes across various frames. For instance, at the top of Fig.\ref{fig:supp_dataset}, while narrating ``loading drugs'', the storyboard displays diverse views, scenes, and characters. Besides, film-sets related to the storyboard and characters are appropriately annotated. Annotations such as speedboat, freighter, magazine, and notebook are provided as well.

\begin{figure}[h]
    \centering
    \includegraphics[width=1\linewidth]{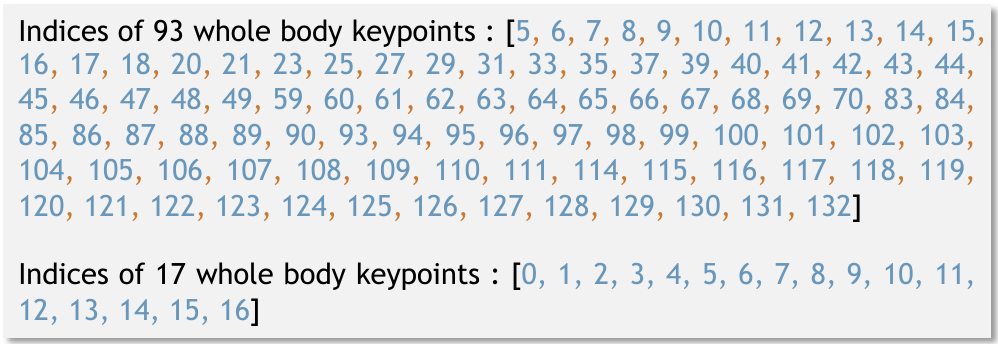}
    \caption{Indices of sampled keypoints.}
    \label{fig:supp_keypoints}
\end{figure}

\section{More implementation Details}

\begin{figure}[h]
    \centering
    \includegraphics[width=1\linewidth]{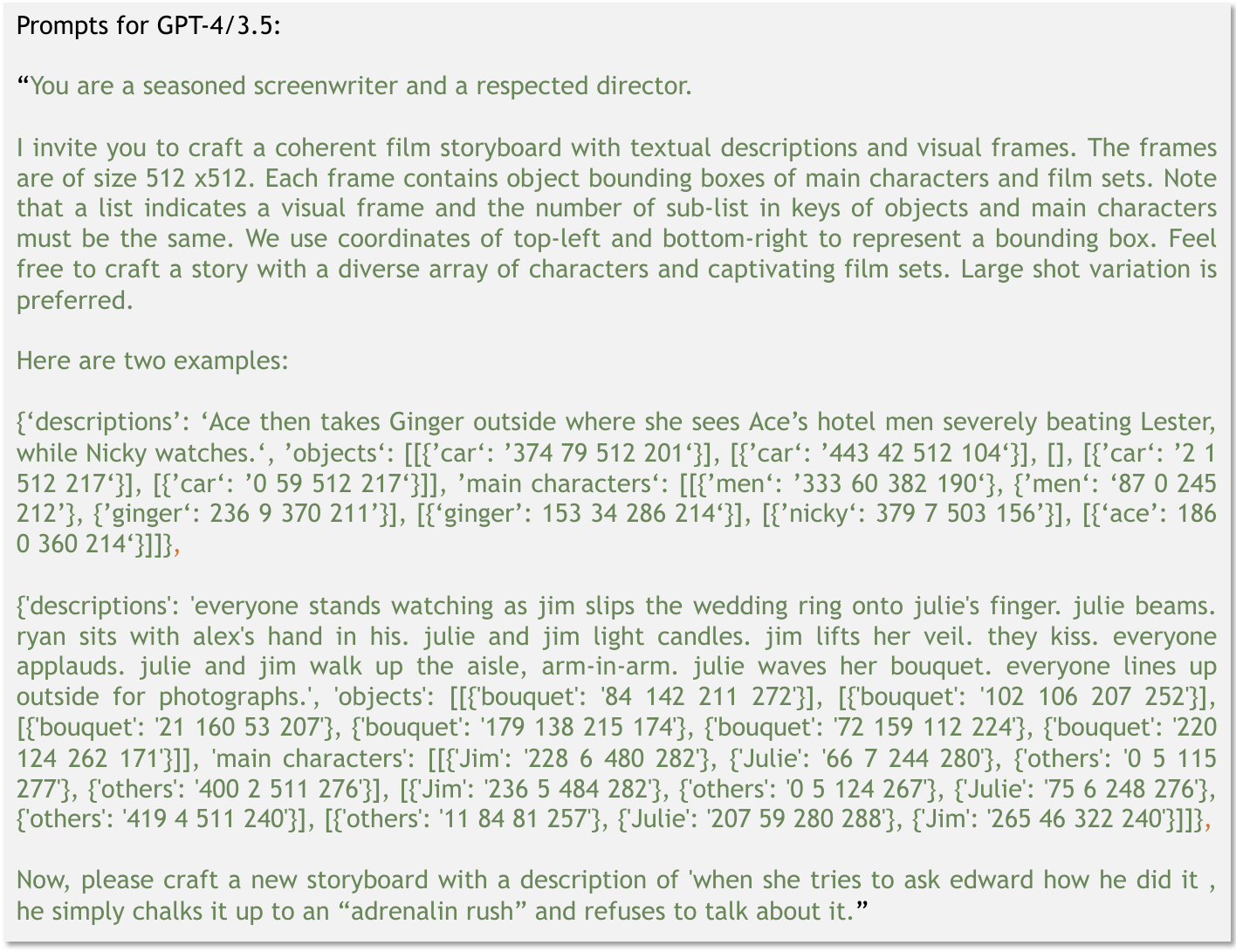}
    \vspace{-10pt}
    \caption{Prompts for GPT-3.5/4.}
    \label{fig:supp_incontext}
    \vspace{-10pt}
\end{figure}

As mentioned earlier, we sample 93 keypoints instead of using all 136 whole-body keypoints. Specifically, when the area of a bounding box falls within the range from $32^2$ to $96^2$, we employ 17 keypoints to represent the characters. Otherwise, we use only the bounding box as the representation of a character. The corresponding indices of these keypoints are listed in Fig.~\ref{fig:supp_keypoints}. We present the prompts for GPT-4/3.5 in Fig.~\ref{fig:supp_incontext}. Specifically, we provide GPT-4/3.5 with two in-context learning examples and prompt it to generate storyboards with large variation across various shots.

\section{Impact of Model Sizes}

\begin{table}[h]
    \centering
    \vspace{-30pt}
     \setlength\tabcolsep{20pt}
    \resizebox{\linewidth}{!}{%
    \begin{tabular}{ccc}
        \toprule
         \# Parameters & FID on \textit{testA} $\downarrow$ & FID on \textit{testB} $\downarrow$ \\
         \midrule
         82.30M / 96.49M / 124.44M & 7.01 / 5.69 / \textbf{4.42} & 33.73 / 30.45 / \textbf{27.09} \\
         \bottomrule
    \end{tabular}}
    \captionsetup[table]{font=scriptsize}
    \caption{The impact of model scaling.}
    \label{tab:modelscale}
    \vspace{-30pt}
\end{table}

We illustrate the impact of increasing the scale of GPT in Table~\ref{tab:modelscale}. As shown in Table~\ref{tab:modelscale}, it is evident that there is a consistent trend of performance improvement as the model size scales up.

\end{document}